\documentclass[letterpaper, 10pt, conference]{ieeeconf}

\IEEEoverridecommandlockouts
\usepackage{cite}
\usepackage{graphicx}
\usepackage{multirow}
\usepackage{array}
\usepackage{booktabs}
\usepackage{makecell}
\usepackage{subcaption}
\usepackage[hidelinks=true]{hyperref}
\usepackage{amsmath}
\usepackage{amssymb}
\usepackage{siunitx}
\usepackage{dblfloatfix}
\usepackage{cuted}
\usepackage{fancyhdr}
\newcommand{\suppurl}{\href{https://tinyurl.com/helix-arm-SI}{\tt\small https://tinyurl.com/helix-arm-SI}}

\title{\LARGE \bf
A Model-Based Decoupling Strategy for Proprioception and Contact Sensing in an Architected Soft Manipulator
}

\author{
  Francesco Stella$^{*1,2,5}$, Annan Zhang$^{*5}$, Cosimo Della Santina$^{3,4}$, Josie Hughes$^{2}$, Daniela Rus$^{5}$
\thanks{$^*$These authors contributed equally.}
\thanks{$^1$ Embodied AI SA, Lausanne, Switzerland}
\thanks{$^2$ CREATE Lab, STI, EPFL, CH-1015 Lausanne, Switzerland}
\thanks{$^3$ Department of Cognitive Robotics, Delft University of Technology, 2628 CD Delft, The Netherlands}
\thanks{$^4$ Institute of Robotics and Mechatronics, German Aerospace Center, 82234 Wessling, Germany}
\thanks{$^5$ MIT Computer Science \& Artificial Intelligence Laboratory, Massachusetts Institute of Technology, Cambridge, MA 02139, USA}
\thanks{Email correspondence to: \href{mailto:zhang@csail.mit.edu}{\tt\small zhang@csail.mit.edu}}
}

\begin{document}

\maketitle
\thispagestyle{fancy}
\fancyhf{}
\renewcommand{\headrulewidth}{0pt}
\fancyhead[C]{\footnotesize Accepted for publication in the proceedings of the \textit{2026 IEEE/RSJ International Conference on Intelligent Robots and Systems (IROS)}}
\fancyfoot[C]{\footnotesize \textcopyright 2026 IEEE. Personal use of this material is permitted. Permission from IEEE must be obtained for all other uses, in any current or future media, including reprinting/republishing this material for advertising or promotional purposes, creating new collective works, for resale or redistribution to servers or lists, or reuse of any copyrighted component of this work in other works.
% DOI: \href{https://doi.org/[add DOI]}{\tt \small [add DOI]}
}
\pagestyle{empty}

%%%%%%%%%%%%%%%%%%%%%%%%%%%%%%%%%%%%%%%%%%%%%%%%%%%%%%%%%%%%%%%%%%%%%%%%%%%%%%%%

\begin{figure*}
  \centering
  \includegraphics[width=0.9\textwidth]{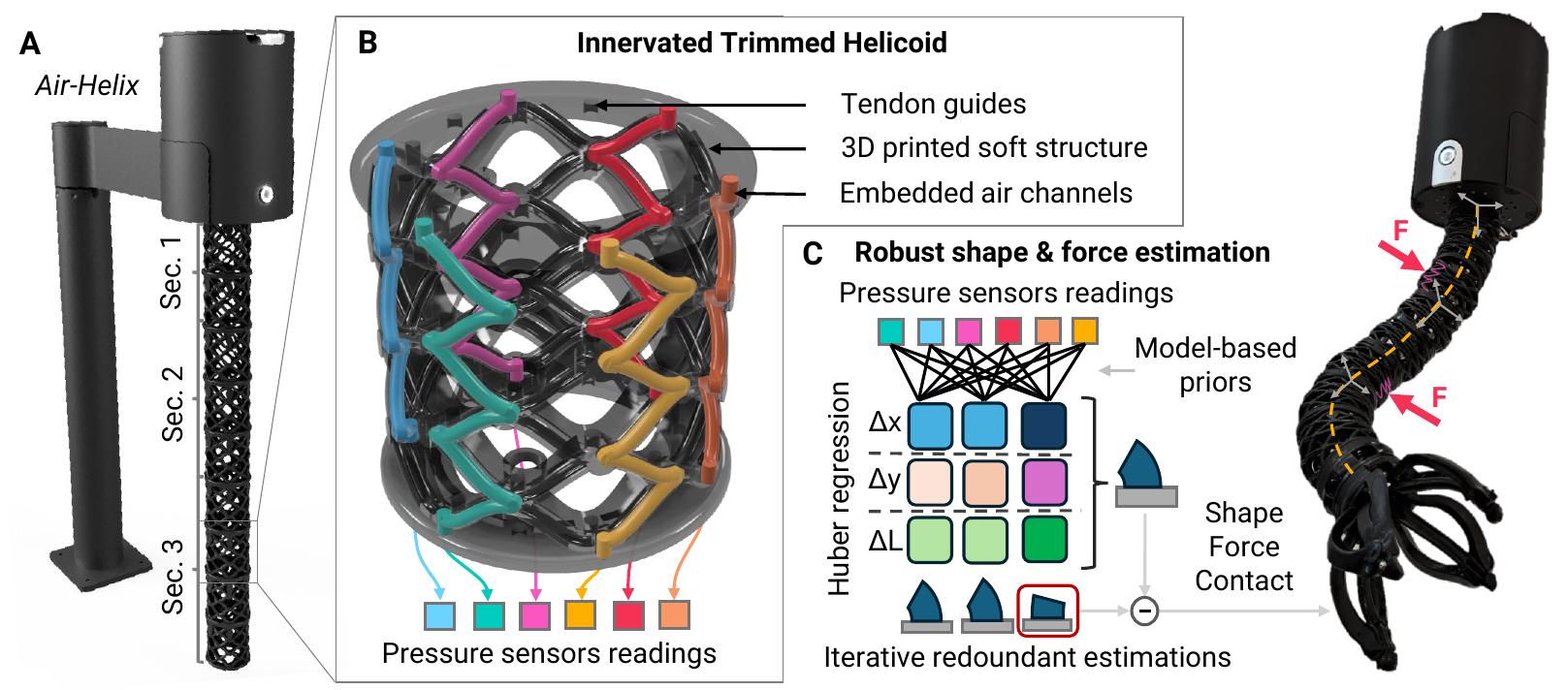}
  \captionof{figure}{\textbf{A} The \textit{Air-Helix} soft manipulator: eight ITH segments form three independently actuated sections, with motors at the base. \textbf{B} Detail of an Innervated Trimmed Helicoid segment with six fluidic channels routed in a localized zigzag pattern along the circumference, connected to pressure sensors. \textbf{C} Decoupling strategy: the six pressure readings are mapped to the three segment DOFs via a PCC model; Huber regression down-weights outlier channels (contact) while estimating shape from the consensus channels.}
  \label{fig:fig1_fluid}
\end{figure*}

\begin{abstract}
Soft continuum robots require embedded sensing for proprioception and contact detection, yet integrating sensors into sparse, highly deformable architected structures remains challenging. We present a model-based strategy that decouples proprioceptive and contact signals from a common set of fluidic pressure sensors embedded in a soft architected segment. Each segment of the Innervated Trimmed Helicoid (ITH) contains six air channels routed in a localized zigzag pattern along the circumference. With only three principal kinematic degrees of freedom (axial compression, bending in $x$, bending in $y$), the six pressure readings form an overdetermined system. A piecewise constant curvature model maps pressures to shape, and Huber regression identifies outlier channels whose residuals indicate external contact. On a single ITH segment, this approach achieves proprioceptive shape estimation with a relative bending error of $0.11 \pm 0.02$ and a contact detection rate of 97\% across 178 trials. We integrate eight ITH segments into \textit{Air-Helix}, a tendon-driven soft continuum manipulator, and present exploratory whole-arm demonstrations that include tactile teaching by demonstration, admittance-controlled force regulation, and tactile object reconstruction. The results suggest that localized fluidic innervation combined with model-based redundancy resolution is a practical path toward concurrent proprioception and contact sensing in architected soft robots.
\end{abstract}

%%%%%%%%%%%%%%%%%%%%%%%%%%%%%%%%%%%%%%%%%%%%%%%%%%%%%%%%%%%%%%%%%%%%%%%%%%%%%%%%
\section{Introduction}

Soft continuum manipulators are valued for their compliance and ability to safely interact with unstructured environments~\cite{rus2015design,della2021soft}. However, the same deformability that makes them attractive also makes sensing difficult: effective closed-loop control and contact-rich manipulation require both proprioception (knowledge of the robot's own shape) and exteroception (detection of external contacts), yet integrating sensors into highly deformable, sparse structures without compromising their mechanical properties remains an open challenge~\cite{hegde2023sensing}.

A variety of embedded sensing modalities have been explored for soft robots. Piezoresistive and capacitive skins can cover large areas but suffer from poor repeatability, parasitic capacitance, and the need for stable mounting surfaces that sparse lattice structures do not provide~\cite{maiolino2013flexible,shorthose2022design,case2025expanded}. Optical fibers~\cite{modes2020shape,cao2022closed} and compact IMUs~\cite{stella2023soft,hughes2021sensing,pei2024imu} offer localized shape estimates but require complex routing and rigid attachments. Vision-based tactile sensors provide high resolution at the end effector, but are bulky and require controlled optical conditions, making whole-body deployment impractical on sparse lattice structures. Liquid metal sensors offer high compliance but face leakage risks and toxicity concerns. Strain gauges and conductive elastomers have been embedded in soft actuators~\cite{truby2020distributed,shih2020electronic}, but strain gauges are fragile under the large strains typical of soft robots, and conductive elastomers suffer from hysteresis, signal drift, and limited durability under repeated loading~\cite{thuruthel2019soft,ang2022learning}. Internal pressure sensors have also been used to reconstruct the 3D deformation of pneumatic soft robots~\cite{scharff2021sensing}, although these approaches are coupled to the pressurized actuators themselves. Across these approaches, most methods measure the robot's shape \emph{or} detect contact, but rarely both through a single sensing modality.

Fluidic innervation, the principle of embedding sealed air channels within a soft structure and reading their internal pressure, has emerged as a sensing approach compatible with 3D-printed architected materials~\cite{truby2022fluidic}. Because the channels are printed as part of the structure itself, they do not require additional mounting surfaces or complex routing, and they preserve the mechanical compliance of the host material. Since its introduction, fluidic innervation has been extended to sensorized grippers that detect contact location and slip~\cite{zhang2024embedded,shang2026forte}, durable tactile sensors~\cite{zhang2025fluidically}, wearable human-robot interfaces~\cite{rubin2026high}, and learned proprioceptive models from soft wrist-like platforms~\cite{zhang2023machine}. Recent work~\cite{zhang2026printed} showed that air channels can be embedded into helicoid-structured segments via multi-material printing, and the helicoid architecture has been further developed for soft-rigid hybrid designs~\cite{patterson2025design}. However, the channels in~\cite{zhang2026printed} follow the helicoid curve continuously over the full segment height, coupling the response of each channel to global deformations in a way that does not naturally separate proprioceptive from contact-induced signals. More broadly, a model-based framework that exploits sensor redundancy to \emph{decouple} proprioception from contact detection using fluidic innervation has not been demonstrated.
A closely related effort by Wang et al.~\cite{wang2024sensing} achieved simultaneous proprioception and contact detection in a rod-driven soft robot by comparing shapes predicted from actuator commands (via a PCC model) against shapes measured by surface strain sensors. Their approach requires knowledge of the actuator inputs and a separate set of surface sensors to detect discrepancies. In contrast, we exploit the geometric redundancy within a single sensor modality: six fluidic channels resolve three kinematic DOFs, and the overdetermined sensory system itself reveals contacts without requiring actuator feedback.

In this work, we present a model-based decoupling strategy for proprioception and contact sensing in an architected soft manipulator segment, and demonstrate its integration in a complete tendon-driven continuum robot.  The key design element is the Innervated Trimmed Helicoid (ITH), an architected structure in which six air channels are routed in a localized zigzag pattern along the circumference of the segment (Fig.~\ref{fig:fig1_fluid}). Unlike continuous helicoid routing in~\cite{zhang2026printed}, this localized topology produces six independent pressure measurements that are analytically related to the three kinematic degrees of freedom of the segment (axial compression and bending in two directions) through a piecewise constant curvature (PCC) model. The resulting overdetermined system (six measurements, three unknowns) enables a Huber regression to simultaneously estimate the segment's shape and flag outlier channels whose residuals indicate localized contact. The contributions of this paper are:
\begin{enumerate}
    \item A localized air channel design for architected helicoid segments that yields six analytically interpretable, redundant pressure measurements per segment;
    \item A model-based decoupling algorithm combining PCC kinematics with Huber regression to jointly estimate shape and detect contacts in real time;
    \item Quantitative single-segment evaluation demonstrating proprioceptive accuracy (relative bending error $0.11 \pm 0.02$) and contact detection reliability (97\% over 178 trials);
    \item Integration into \textit{Air-Helix}, an eight-segment tendon-driven manipulator, with exploratory whole-arm demonstrations of tactile teaching and tactile object reconstruction.
\end{enumerate}

\section{Methods}

\subsection{Innervated Trimmed Helicoid Design}

In this work we expand the design of the Trimmed Helicoid, a 3D-printed architected structure based on helicoid geometry, first introduced for soft continuum robots in\cite{guan2023trimmed}. This original design is parametrically designed to balance axial and bending stiffness for a wide workspace and homogeneous compliance. The Innervated Trimmed Helicois (ITH) proposed in this article extends this design by embedding six independent air channels along the circumference, spaced $60^\circ$ apart  (Fig.~\ref{fig:fig1_fluid}B).

A key design choice distinguishes the channel routing in this work from prior fluidic innervation of helicoid structures~\cite{zhang2026printed}. Rather than routing the channels continuously along the helicoid curve over the full segment height, each channel here follows a localized zigzag path. This localized topology ensures that each channel's pressure response is predominantly governed by the local deformation at its circumferential position, rather than being averaged over the full segment. The resulting six independent pressure signals are analytically related to the three segment DOFs through the kinematic model described in Sec.~\ref{sec:sensor_model}.

Each channel has an internal diameter of 3\,mm and an external diameter of 4\,mm, selected through a fabricability study (see supplementary material\footnote{Available at \suppurl}, Fig.~S1). The channel follows five zigzag turns along the segment height of approximately 65\,mm, for a total channel length of approximately 130\,mm. The channels are sealed at one end and connected to differential pressure sensors (All Sensors) at the base via 1\,mm rubber tubing.

The sensing principle follows from the ideal gas law: each sealed channel encloses a fixed mass of air, so when the structure deforms, the local change in channel cross-section alters the enclosed volume and produces a pressure change at the differential sensor. Because air is compressible, the response integrates deformation over the full channel length. The signal is therefore uniform regardless of where along the channel the deformation occurs. This is a key advantage over resistive or capacitive point sensors. A second, homologous chamber on the sensor's reference port compensates for temperature-induced drift. Details of the 3D printing (DLP on Carbon printers, EPU~40) and channel cleaning are in the supplementary material.

\subsection{Sensor-to-Shape Model}
\label{sec:sensor_model}

We model each ITH segment as a piecewise constant curvature (PCC) section~\cite{della2020improved,webster2010design}, fully described by three configuration variables $q = [\Delta L, \Delta x, \Delta y]^\top$ representing changes in central length and bending in two orthogonal directions. The length of the $j$-th channel is related to the segment configuration by
\begin{equation}
L_j = L_0 + \Delta L + \Delta x \cos\alpha_j + \Delta y \sin\alpha_j
\label{eq:hmapping}
\end{equation}
where $L_0$ is the undeformed segment length and $\alpha_j \in \{0, \frac{\pi}{3}, \frac{2\pi}{3}, \pi, \frac{4\pi}{3}, \frac{5\pi}{3}\}$ is the angular position of the $j$-th channel. We measure each channel length $\hat{L}_j$ from its own pressure reading $p_j$. A per-channel linear calibration $\hat{L}_j = L_0 + s_j\,p_j$ is fit to the Instron force-displacement-pressure curves (Sec.~\ref{sec:characterization}). This measurement is sensory and needs no tendon or actuator feedback.

Writing the six channel measurements as $b = [\hat{L}_1, \ldots, \hat{L}_6]^\top$, the system takes the form $b = Aq$ with
\begin{equation}
A =
\begin{bmatrix}
1 & \cos\alpha_1 & \sin\alpha_1 \\
1 & \cos\alpha_2 & \sin\alpha_2 \\
\vdots & \vdots & \vdots \\
1 & \cos\alpha_6 & \sin\alpha_6
\end{bmatrix}
\in \mathbb{R}^{6 \times 3}
\end{equation}
This system is overdetermined: six measurements for three unknowns. The ordinary least-squares solution $\hat{q} = (A^\top A)^{-1} A^\top b$ provides the shape estimate in the absence of contact. The standard PCC forward kinematics then maps $q$ to end-effector pose (see supplementary material for full expressions). For the full \textit{Air-Helix} robot, the shape of each segment is estimated independently, and the segment-to-segment homogeneous transformations are composed sequentially to obtain the full-arm pose. The tendon-to-shape mapping $q = h_q(\sigma)$, which relates motor displacements $\sigma$ to configuration variables, is used during Cartesian teleoperation (see supplementary material).

\subsection{Huber Regression for Joint Shape and Contact Estimation}
\label{sec:huber}

When an external contact deforms one or more channels beyond the PCC prediction, the corresponding measurements become outliers in the least-squares fit. We exploit this by replacing ordinary least squares with Huber regression via Iterative Reweighted Least Squares (IRLS), which simultaneously estimates the segment shape and identifies contact channels.
The residual for the $j$-th channel is
\begin{equation}
r_j = \hat{L}_j - (L_0 + \Delta L + \Delta x \cos\alpha_j + \Delta y \sin\alpha_j)
\label{eq:residual}
\end{equation}
The Huber loss down-weights large residuals:
\[
\rho(r_j) =
\begin{cases}
\frac{1}{2} r_j^2 & \text{if } |r_j| \leq \delta \\
\delta |r_j| - \frac{1}{2}\delta^2 & \text{if } |r_j| > \delta
\end{cases}
\]
where $\delta = 1.5 \cdot \mathrm{MAD}$, with MAD the median absolute deviation computed from a calibration sequence of random deformations without contact (cf.\ Fig.~\ref{fig:proprioception_fluid}A). The IRLS procedure is:
\begin{enumerate}
    \item Initialize $\hat{q}$ from the ordinary least-squares solution.
    \item Compute residuals $r_j$ and weights $w_j = 1$ if $|r_j| \leq \delta$, else $w_j = \delta / |r_j|$.
    \item Solve $\hat{q} = (A^\top W A)^{-1} A^\top W b$ with $W = \mathrm{diag}(w_1, \ldots, w_6)$.
    \item Repeat steps 2--3 until convergence.
\end{enumerate}
After convergence, channels with $w_j < 1$ are flagged as contacts. The algorithm converges in 0.2\,ms (MATLAB), and enables real-time operation at the 100\,Hz sensor rate. This step enables joint shape and contact reconstruction, as shown in Section \ref{results}.

\begin{figure*}%[t!]
    \centering
    \includegraphics[width=0.95\textwidth]{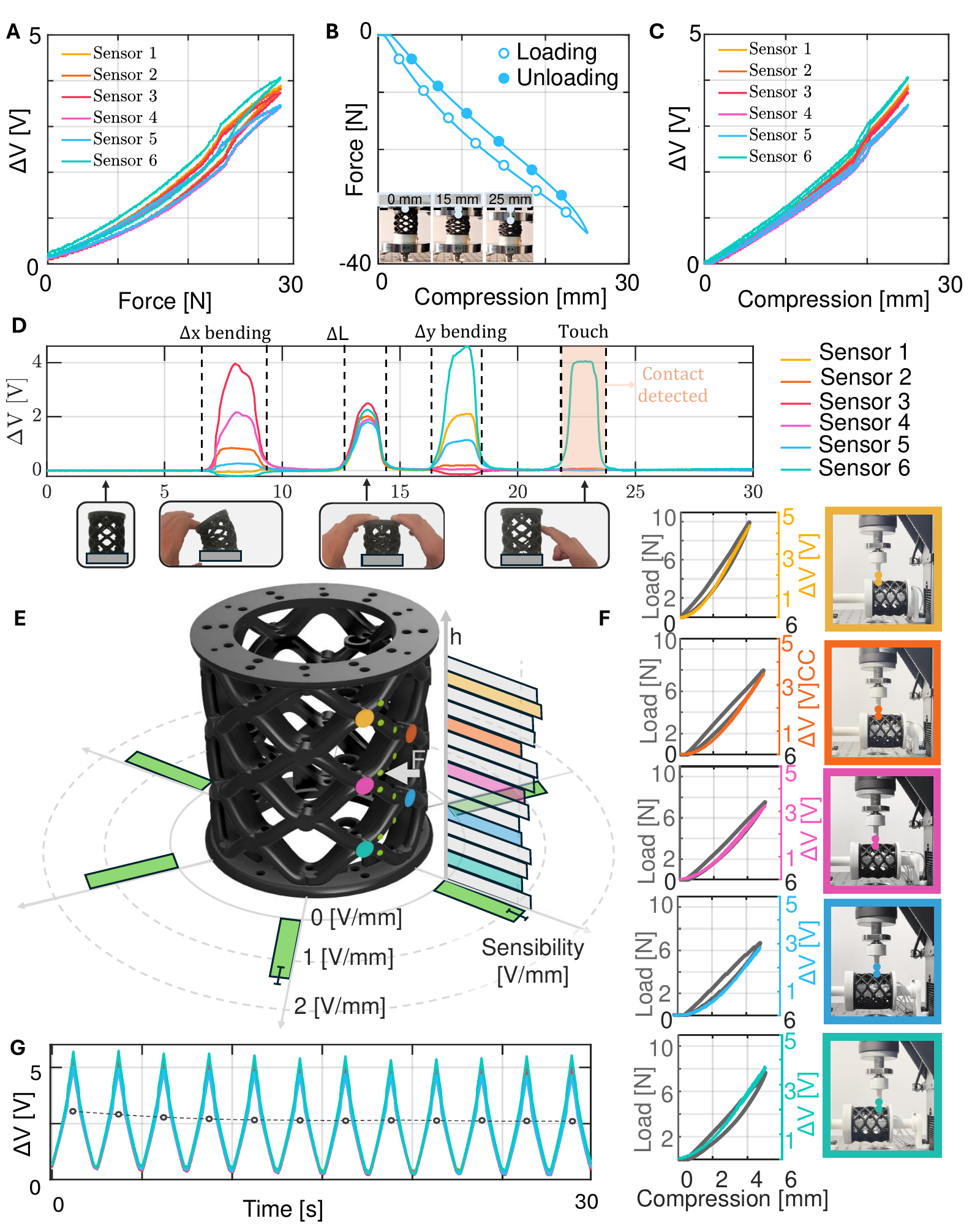}
    \caption{Single-segment characterization. \textbf{A--C)} Force, compression, and pressure sensor readings during Instron loading. The approximately linear pressure-compression relationship enables the calibration mapping used in (\ref{eq:hmapping}). \textbf{D)} Raw readings during manual deformation: bending in $x$, compression, bending in $y$, and localized touch produce distinguishable sensor signatures. \textbf{E--F)} Sensitivity analysis across 15 radial positions per channel and 5 loading levels, showing coherent response along and across channels. \textbf{G)} Cyclic Instron compression showing repeatable sensor response after 4 preconditioning cycles.}
    \label{fig:charact_fluid}
\end{figure*}

\section{Results}
\label{results}
We evaluate the proposed decoupling strategy at the single-segment level with quantitative ground-truth comparison, and then present exploratory demonstrations on the full \textit{Air-Helix} manipulator.

\subsection{Single-Segment Characterization}
\label{sec:characterization}

We characterize a single ITH segment under controlled loading using an Instron machine (Fig.~\ref{fig:charact_fluid}). The segment is axially compressed while recording both force-displacement and pressure sensor data simultaneously.

The material (EPU~40) exhibits repeatable behavior after four preconditioning cycles (Fig.~\ref{fig:charact_fluid}G), consistent with prior durability studies~\cite{stella2024toward}. The pressure response of all six channels is approximately linear with compression (Fig.~\ref{fig:charact_fluid}A--C), which justifies the linear calibration mapping used in~(\ref{eq:hmapping}). Linear interpolation of these curves yields the per-channel pressure-to-length mapping $\hat{L}_j(p_j)$ used in Sec.~\ref{sec:sensor_model}.

To assess spatial uniformity, we performed a sensitivity analysis by applying controlled forces at 15 radial positions per channel (Fig.~\ref{fig:charact_fluid}E--F). The response is coherent across positions along each channel, confirming that the enclosed air volume integrates local deformation uniformly over the full channel length. Across channels, the sensitivity varies moderately due to manufacturing tolerances in the DLP printing process, but the per-channel linear calibration compensates for these differences. Manual deformation tests (Fig.~\ref{fig:charact_fluid}D) confirm that bending in $x$, axial compression, bending in $y$, and localized touch each produce distinguishable sensor signatures, validating that the six-channel arrangement can resolve these distinct deformation modes.

\begin{figure*}[t!]
    \centering
    \includegraphics[width=0.9\textwidth]{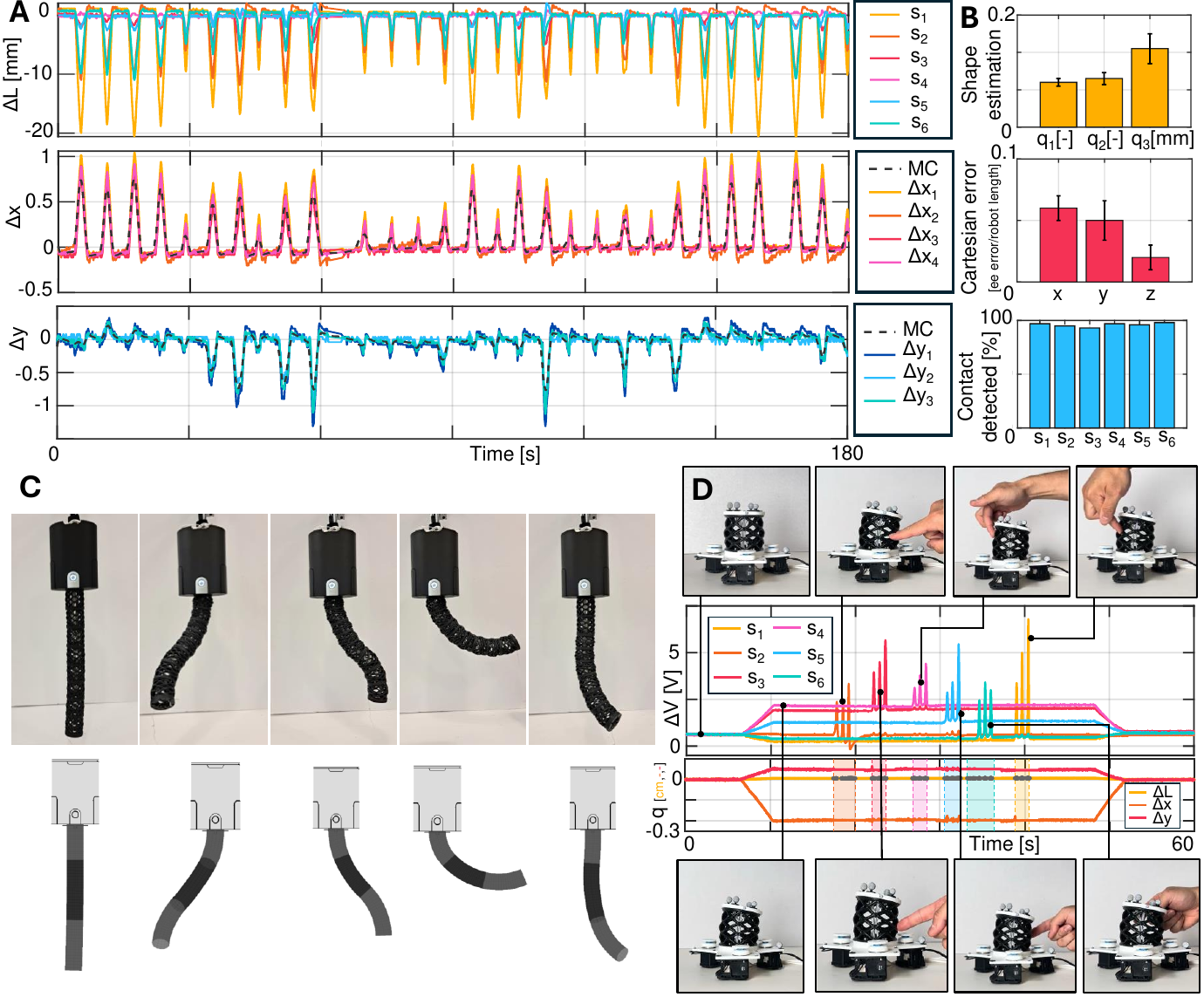}
    \caption{Single-segment proprioception and contact detection. \textbf{A)} Time series of estimated shape variables ($\Delta L$, $\Delta x$, $\Delta y$) from redundant sensor subsets, compared to OptiTrack ground truth during random tendon-driven motions. \textbf{B)} Summary metrics: single-section shape error, whole-robot end-effector Cartesian error, and contact detection rate. \textbf{C)} Full robot pose reconstructed from embedded sensing (bottom) vs.\ photographs (top). \textbf{D)} Contact detection during a fixed-pose experiment: sensor signals (top) and the Huber regression shape estimate (bottom) remain stable while six sequential touches are correctly flagged.}
    \label{fig:proprioception_fluid}
\end{figure*}

\begin{figure*}[t!]
    \centering
    \includegraphics[width=0.82\textwidth]{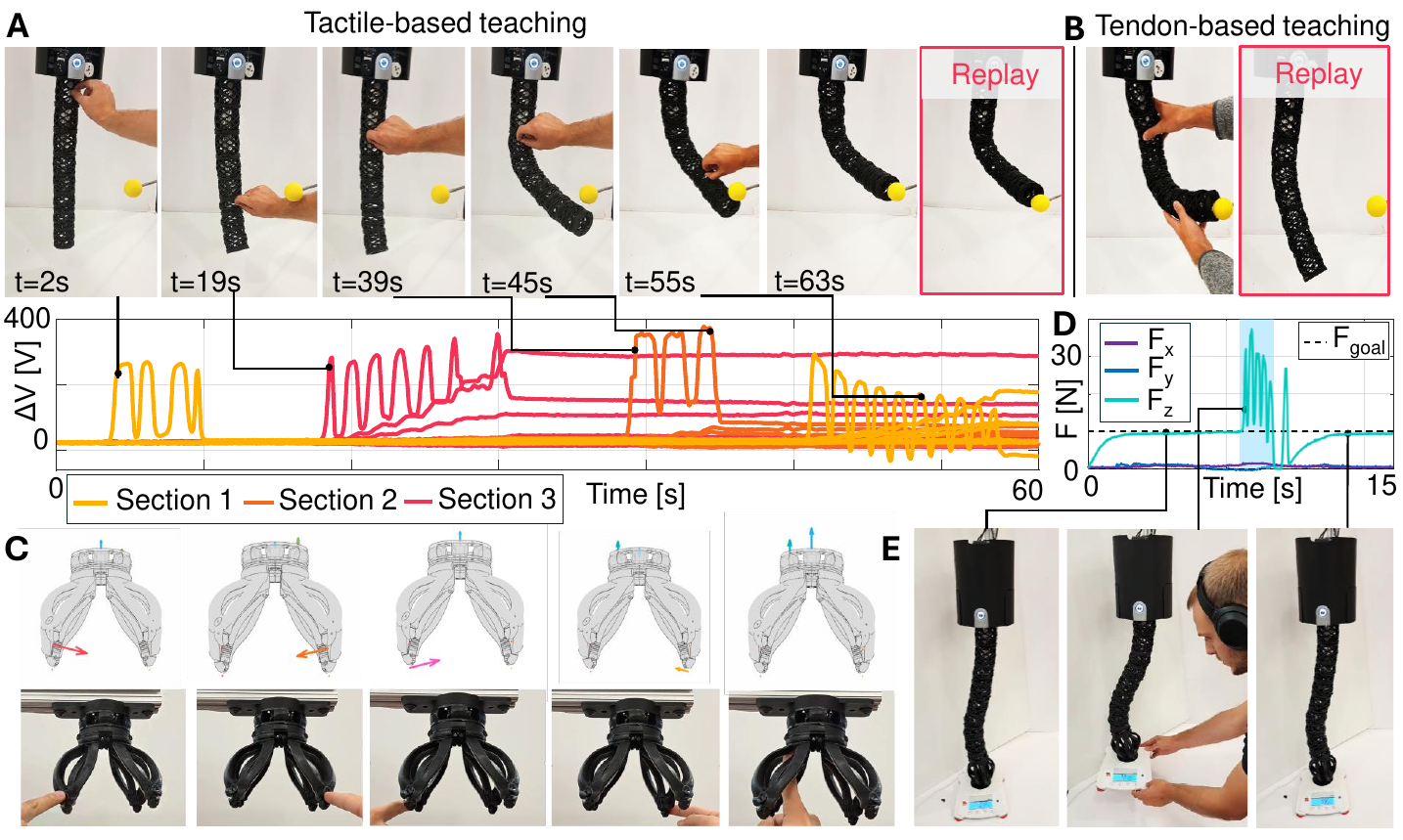}
    \caption{Exploratory whole-arm demonstrations. \textbf{A)} Tactile teaching: the user presses the robot body to indicate desired bending locations; contacts trigger local tendon contractions, building up the target shape incrementally. Replay is accurate. \textbf{B)} Baseline tendon-based teaching: the user physically deforms the robot; replay is inaccurate due to tendon transmission losses. \textbf{C)} Force estimation at the gripper via redundant air chambers. \textbf{D--E)} Admittance control maintaining a target force at the end effector.}
    \label{fig:fluid_gripper}
\end{figure*}

\subsection{Proprioceptive Accuracy}
\label{sec:proprioception}

A single ITH segment was actuated by three tendons executing random, non-periodic motions while an OptiTrack system recorded ground-truth tip position. The sensor-to-shape model (Sec.~\ref{sec:sensor_model}) and Huber regression (Sec.~\ref{sec:huber}) were applied in real time to estimate the segment state.

Fig.~\ref{fig:proprioception_fluid}A shows a representative time series of the estimated $\Delta x$ and $\Delta y$ compared to ground truth. The redundant estimates from different sensor subsets are mutually consistent and track the motion-capture reference at high frequency. Over the full dataset (average actuation speed 1\,rad/s), the relative bending error was $0.11 \pm 0.02$ (mean $\pm$ std). The supplementary material defines all reported error metrics.

The same algorithm was deployed on the full eight-segment \textit{Air-Helix} robot. Fig.~\ref{fig:proprioception_fluid}C shows qualitative agreement between reconstructed and actual poses. The end-effector Cartesian error, normalized by total robot length, was $0.06 \pm 0.02$ (Fig.~\ref{fig:proprioception_fluid}B). We note that the whole-arm evaluation used static poses rather than continuous motion capture, so it should be interpreted as a preliminary assessment (Fig.~\ref{fig:proprioception_fluid}C).

\subsection{Contact Detection}
\label{sec:contact}

To evaluate the decoupling of proprioception and contact, a single segment was actuated to a fixed configuration, and an operator sequentially touched each of the six channel locations on the segment surface. Fig.~\ref{fig:proprioception_fluid}D shows that the Huber regression shape estimate ($\Delta L$, $\Delta x$, $\Delta y$) remains stable throughout the contact sequence, while the raw sensor signals clearly register each touch event. Channels with residuals exceeding the Huber threshold $\delta$ are flagged as contacts.

The detection threshold $\delta = 1.5 \cdot \mathrm{MAD}$ was set using a 60-second calibration phase during which the segment was actuated through random configurations without external contact. This calibration captures the natural residual variability due to model mismatch and sensor noise, establishing the baseline above which a residual is treated as contact-induced. The threshold can be adjusted to trade off sensitivity against false-positive rate; the value $1.5 \cdot \mathrm{MAD}$ was found empirically to yield a good balance for the contact forces applied in our experiments.

This experiment was repeated across 9 fixed configurations uniformly distributed over the workspace, totaling 178 contact events. The mean contact detection rate was 97\% across all six channels (Fig.~\ref{fig:proprioception_fluid}B). The spatial resolution of contact localization is limited to $60^\circ$ sectors (one per channel); contacts falling between channels activate the nearest one. A single channel cannot localize contact along its own length since the pressure is uniform across the channel volume. Instead, the system compares residuals across channels and segments. This locates each contact to one $60^\circ$ sector and one segment, about 65\,mm axially. The current evaluation is limited to single contacts at a time.

\begin{figure*}[t!]
    \centering
    \includegraphics[width=0.82\textwidth]{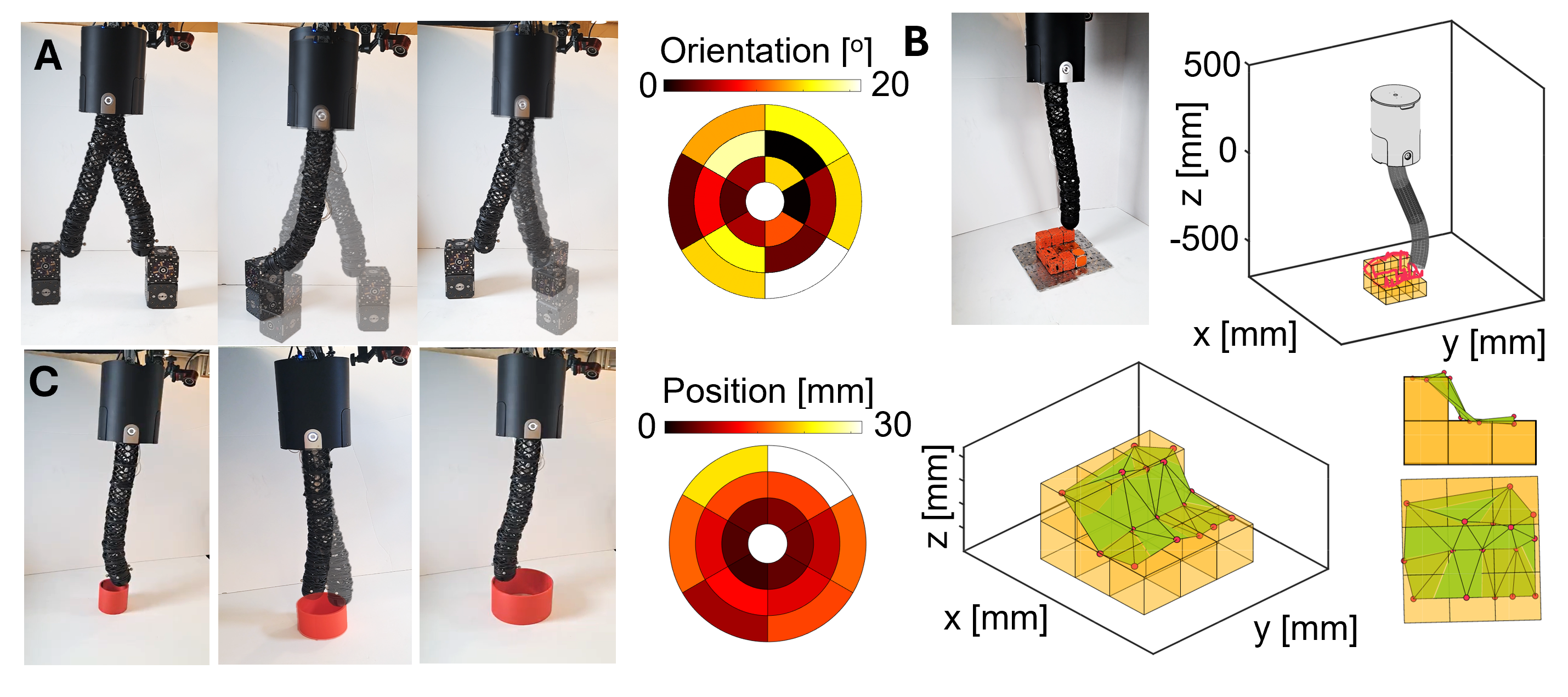}
    \caption{Tactile object reconstruction. \textbf{A)} The robot is teleoperated to contact objects at known locations; bull's-eye plots show position and orientation estimation accuracy, which degrades at larger deformations. \textbf{B)} Contact point cloud from repeated interactions with an unknown object, reconstructed via Delaunay triangulation. \textbf{C)} Cylinder radius estimation from contact points (accuracy $\pm 1.6$\,cm).}
    \label{fig:SLAM}
\end{figure*}

\section{Exploratory Whole-Arm Demonstrations}

The single-segment results above establish the decoupling capability under controlled conditions. In this section, we present three exploratory demonstrations on the full \textit{Air-Helix} manipulator that illustrate how the joint proprioceptive and contact-sensing capabilities could be used in practice. These demonstrations are qualitative and should be understood as proofs of concept; rigorous quantitative evaluation of the full multi-segment system is left for future work.

\subsection{Tactile Teaching by Demonstration}

Teaching by demonstration on soft continuum arms is challenging because underactuated tendon-driven systems do not directly measure body deformation, and tendon transmission losses cause significant replay error. A baseline tendon-based teaching approach (Fig.~\ref{fig:fluid_gripper}B), in which the user physically deforms the robot with constant tendon tension and motor positions are recorded, produces noticeable error during replay.

The decoupling strategy enables an alternative: tactile teaching (Fig.~\ref{fig:fluid_gripper}A). The user presses the robot body at the location where bending is desired. The contact detection algorithm identifies the pressed segment and channel, and the nearest tendon is contracted by a fixed increment $\Delta\sigma$. By repeating this process along the body, the user incrementally builds the desired configuration using only internal actuation forces. Since the final shape is achieved through tendon positions rather than external deformation, replay is accurate.

\subsection{Force Sensing and Admittance Control}

The Instron characterization (Fig.~\ref{fig:charact_fluid}) provides a mapping from channel deformation to applied force. This mapping allows estimating contact force magnitude from the Huber regression residuals: a channel flagged as an outlier has a residual $r_j$ corresponding to excess deformation beyond the PCC model prediction, which can be translated into a force estimate via the calibration curve.

To explore this capability at the end effector, we designed and fabricated a gripper innervated with multiple air chambers: six at the connection with the arm and four in the gripper's fingers (Fig.~\ref{fig:fluid_gripper}C; see supplementary material for design details). By combining the local force readings at the connection level, the forces and torques at the end effector along $[x,y,z]$ axes are estimated.

Using this force estimate, a proportional admittance controller was implemented to maintain a target force at the end effector (see supplementary material for the control law). As shown in Fig.~\ref{fig:fluid_gripper}D--E, the robot was controlled in real time to maintain a desired force profile: when external perturbations increase the contact force, the robot retracts to restore the target, and vice versa. This demonstrates that the fluidic sensing can close the loop on force-controlled interaction.

\subsection{Tactile Object Reconstruction}

Combining proprioception with end-effector contact detection, the robot can estimate the location of objects in its workspace. The robot was teleoperated in Cartesian space (see supplementary material) with a hemispherical sensorized fingertip (supplementary material, Fig.~S3) replacing the gripper. Contact events were detected via a pressure threshold on the tip's embedded air chambers.

In a first experiment (Fig.~\ref{fig:SLAM}A), the robot contacted objects at known positions to evaluate localization accuracy against motion capture. Accuracy degrades with distance from the base due to accumulated shape estimation error along the kinematic chain and higher sensitivity to external forces due to a larger moment arm from the base of the robot. The robot also estimated the radius of three cylinders with an accuracy of $\pm 1.6$\,cm (Fig.~\ref{fig:SLAM}C), and reconstructed the surface of an unknown object from a contact point cloud using Delaunay triangulation (Fig.~\ref{fig:SLAM}B). These experiments relied on end-effector contact only and did not exploit whole-body contact detection; extending the approach to leverage contacts along the full arm is a direction for future work.

\section{Discussion and Limitations}

We presented a model-based strategy for decoupling proprioceptive and contact signals from redundant fluidic pressure sensors embedded in an architected soft manipulator segment. The approach relies on two elements: (i) a localized zigzag channel design that produces six analytically interpretable pressure measurements per segment, and (ii) a Huber regression that separates shape-consistent signals from contact-induced outliers. At the single-segment level, the method achieves a relative bending error of $0.11 \pm 0.02$ and a 97\% contact detection rate.

Several limitations should be noted. First, the decoupling was validated on a single segment with single-point contacts; simultaneous multi-point contacts and the interaction between contacts on adjacent segments were not tested. With six sensors and three DOFs per segment, up to three contacts per segment are theoretically detectable, but this limit has not been experimentally established. Second, the contact detection provides localization only at the resolution of individual channels ($60^\circ$ sectors), not continuous spatial estimation. Third, the linear pressure-to-length calibration relies on the Instron characterization at a fixed temperature and loading rate; the effects of dynamic loading, and long-term drift were not systematically studied. The supplementary material discusses the robustness of the pressure model further. Fourth, the whole-arm demonstrations are exploratory: the full-arm proprioceptive accuracy was assessed against static poses rather than continuous motion capture, and the tactile teaching and object reconstruction experiments were not evaluated quantitatively.

However, compared to the expectation-based approach of Wang et al.~\cite{wang2024sensing}, the method presented here does not require actuator command feedback or a model mapping commands to expected shape, and uses a single sensor modality rather than combining actuator feedback with separate surface sensors. This simplifies deployment on tendon-driven systems, where actuator-to-shape models are inaccurate due to friction and cable losses. Moreover, this same sensory approach could be combined with methods to learn the sensor-to-state mapping directly from data, as explored for fluidic innervation in~\cite{zhang2023machine}, which may be beneficial in multi-contact scenarios. While learned models can capture nonlinearities that the PCC model neglects, they require training data for each robot configuration and do not naturally separate proprioceptive from contact signals without explicit supervision. Our model-based approach has the advantage of providing interpretable residuals that directly indicate contact location, without task-specific training. Relative to prior work on fluidic innervation of helicoid segments~\cite{zhang2026printed}, this paper's contribution is the localized channel design and the model-based decoupling algorithm that exploits the resulting redundancy, rather than the fabrication method itself.

Future work should address multi-contact detection limits and coupled contacts between adjacent segments through systematic testing, improve spatial resolution through denser channel layouts or signal interpolation, and validate the full-arm decoupling capability with continuous ground-truth tracking during dynamic contact scenarios.

\section*{Acknowledgments}

This work was supported in part by the Singapore MIT Alliance on Research and Technology (SMART) under the Mens, Manus, et Machina (M3S) program, by the BARI EMERGE program under grant N00014-26-1-2304, and by the MIT-GIST collaboration. C.D.S.\ acknowledges support from the European Research Council Project RIPLEy. The authors thank Embodied AI SA for providing technical support and the motorized base for the \textit{Air-Helix} robot, and Miguel Flores-Acton and Andy Yu for help with 3D printing, PCB design, and mechanical design.

\bibliographystyle{IEEEtran}
\bibliography{scibib}

\end{document}